\documentclass[]{article}

\usepackage{graphicx}
\usepackage{subcaption}
\title{Automated univariate time series forecasting with regression trees}
\author{Francisco Mart\'inez \\ 
	Department of Computer Science, University of Ja\'en 
	\and 
	Mar\'{i}a Pilar Fr\'ias \\ 
	Department of Statistics and Operations Research, University of Ja\'en}

\begin{document}

\maketitle

\begin{abstract}
This paper describes a methodology for automated univariate time series forecasting using regression trees and their ensembles: bagging and random forests. The key aspects that are addressed are: the use of an autoregressive approach and recursive forecasts, how to select the autoregressive features, how to deal with trending series and how to cope with seasonal behavior. Experimental results show a forecast accuracy comparable with well-established statistical models such as exponential smoothing or ARIMA. Furthermore, a publicly available software implementing all the proposed strategies has been developed and is described in the paper.
\end{abstract}

\section{Introduction}
\label{sec_intro}

Time series forecasting is a crucial activity in many areas, such as sales, health, energy \cite{HUSSEIN2024} or human resources, to name a few. In this context, it is always useful to have automatic tools, especially in some fields, e.g. retail sales, in which a great numbers of series must be forecast, often in a short period of time. 

A way to facilitate automatic prediction is to apply an univariate time series forecast approach \cite{fpp3}, i.e.,  use only the values in the series to predict its future behavior. Thus, there is no need to determine, collect and incorporate external variables into the forecasting process. Another way to make automatic forecasting easier is to avoid parameter tuning. Although parameter tuning can be automated \cite{Charte2020}, it might have a high computational cost and, if the time series are short, it can produce poor results because small training sets might generate  underfitted models and small validation sets may not be useful to correctly estimate the forecasting accuracy of a model.

Regression trees \cite{Breiman:1984} and especially bagging \cite{Breiman:1996} and random forests \cite{Breiman:2001} have interesting properties for an automated time series forecasting tool. They can be used without applying a careful feature selection strategy because of their automatic feature selection capabilities. Furthermore, in the case of bagging and random forests, they can be used without tuning their parameters because they tend to work well with default values. This paper develops a methodology for automated univariate time series forecasting using regression trees bagging and random forests. The use of gradient boosting machines \cite{friedman2001}, another way of combining regression trees, has not been contemplated, because gradient boosting machines require a precise tuning. 

This paper starts with a concise description of regression trees in Section \ref{sec_rt}. Then, Section \ref{sec_rt_utsf}, explains how a general machine learning regression model, and in particular regression trees, can be applied to univariate time series forecasting using an autoregressive approach and recursive predictions. Section \ref{sec_bagging_rf} describes briefly bagging and random forests. Regression trees cannot properly forecast trending series, in Section \ref{sec_trend} three strategies to deal with trending series are discussed, two of the strategies are original proposals. Another key aspect for effectively forecasting a time series is to capture its potential seasonal behavior, Section \ref{sec_seasonality} explains how, carefully selecting the autoregressive features, seasonal series can be suitably forecast. In Section \ref{sec_proposal} our proposal for automated forecasting of univariate time series is described. All the different strategies and tools suggested in the paper have been implemented in an R package, available through CRAN, the official repository of R packages. This package is succinctly discussed in Section \ref{sec_software}. Finally, Section \ref{sec_conclusions} draws some conclusions.

\section{Regression trees}
\label{sec_rt}

A regression tree \cite{Breiman:1984} is a non-parametric algorithm that recursively partitions the feature space into smaller non-overlapping regions using a splitting criterion that tries to find regions with similar response values. Regression trees use binary partitions and the splits are based on yes-or-no questions about the values of a feature. The prediction associated with a region is a constant: the average response value of the examples falling in that region. Binary splits are done in a local, greedy way: the best feature, and splitting point for that feature, is selected so that the overall sum of squared errors between the actual responses and the predicted constants associated with the sub-regions is minimized. The binary recursive partition process goes on until a stopping criterion is reached, e.g., all the examples in a region contain the same response value or a maximum depth is reached.

\begin{figure}
	\centering
	\includegraphics[scale=0.6]{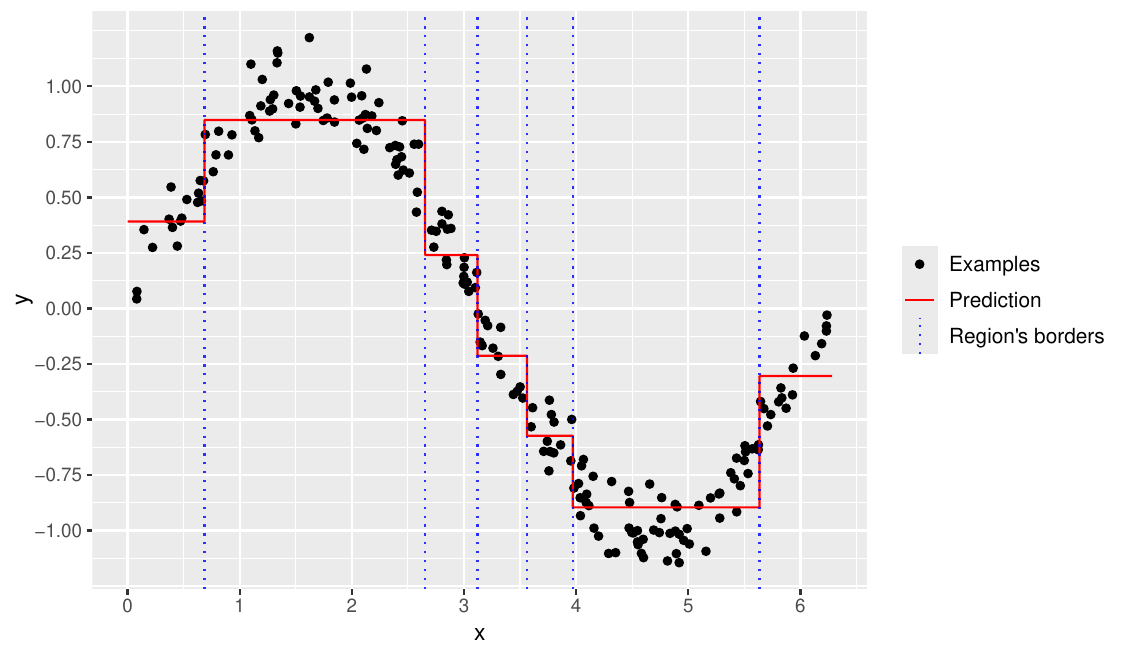}
	\caption{Feature space partition using a regression tree.}
	\label{fig_fsp}
\end{figure}

\begin{figure}
	\centering
    \includegraphics[scale=0.6]{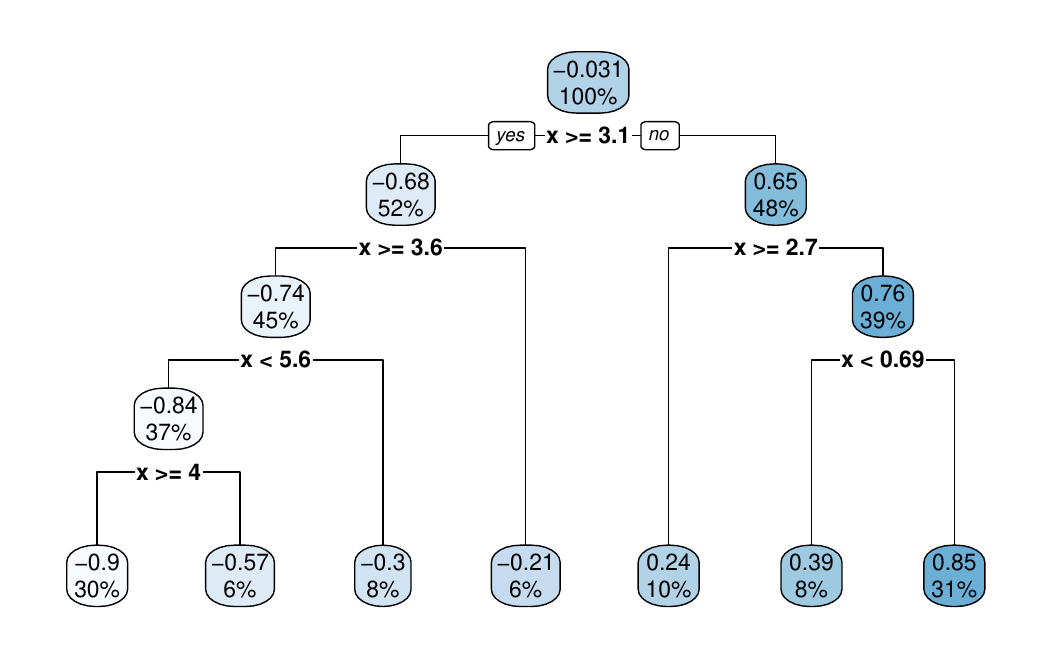}
    \caption{Regression tree associated with feature space subdivision in Fig. \ref{fig_fsp}.}
    \label{fig_rt1}
\end{figure}

Let us see an example in which the training set has only one feature. Figure \ref{fig_fsp} shows a training set with one feature ($x$) and one response ($y$). After applying the regression tree algorithm the feature space is partitioned into the regions delimited by vertical dotted blue lines. The red line is the prediction, consisting of the average response value of the examples falling into a region. The regression tree in Figure \ref{fig_rt1} stores how the feature space is recursively divided. Every node in Figure \ref{fig_rt1} represents a region of the feature space and contains the percentage of training examples falling in that region and its mean response value. For example, the root node represents the whole feature space with 100\% training examples and a mean response value of -0.031. The root node and the internal nodes also show the feature and splitting point at which their associated region is binary subdivided. For example, the root node indicates that the first recursive division was done at $x = 3.1$. The leaves of the regression tree correspond to the final regions of the space partition. As the training set has only one feature, the boundaries of the regions are lines parallel to the y-axis.

To obtain a prediction for an input point, its features are used to consult the yes-or-no questions of the regression tree and find the region of the feature space partition in which the point falls in: the average response value of that region is the prediction for the input point. For instance, given the previous example regression tree and an input point $x = 2$, the point falls in the second region from the left in Figure \ref{fig_fsp}. The prediction for the input point is 0.85, the mean response value of the training examples in that region, as can be seen from the regression tree in Figure \ref{fig_rt1}. 

\begin{figure}
\centering
	\begin{subfigure}{0.49\textwidth}
		\includegraphics[width=\textwidth]{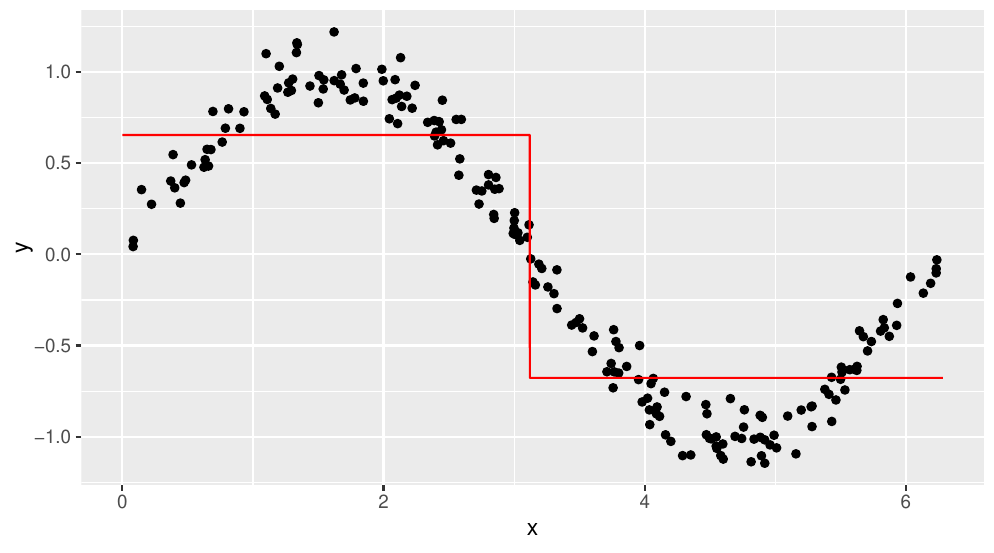}
	    \caption{Underfitted.}
	\end{subfigure}
    \hfill
	\begin{subfigure}{0.49\textwidth}
	\includegraphics[width=\textwidth]{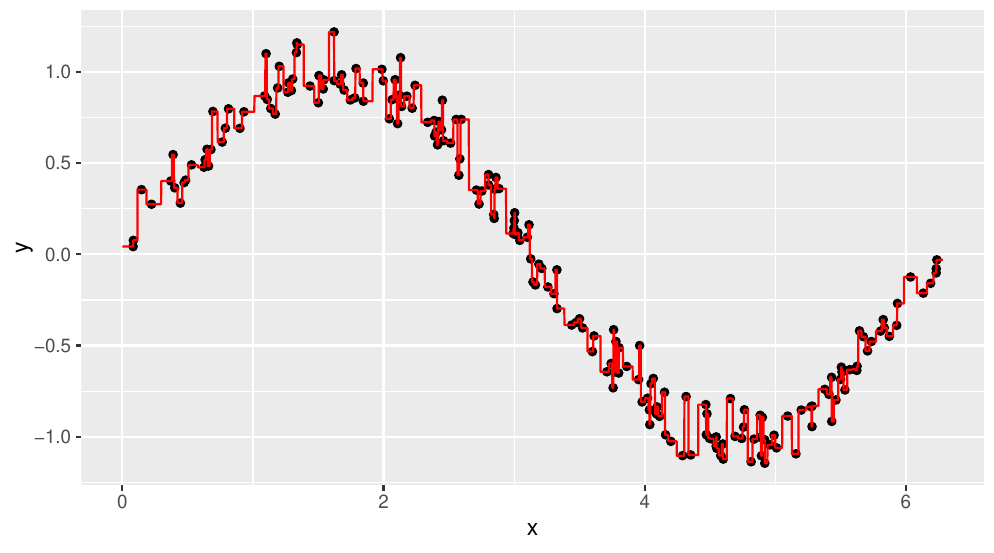}
	\caption{Overfitted.}
\end{subfigure}
    \caption{Underfitted and overfitted regression tree models for the example data set.}
    \label{fig_under_over}
\end{figure}

The depth of the regression tree is key to achieve a good balance between bias and variance in fitting the training set. For example, Figure \ref{fig_under_over} shows the predictions for two regression trees fitted to the previous data set. The left regression tree has depth one and underfits the data set. On the other hand, the right regression tree partitions the feature space too much, so that every final region contains only one example, overfitting the data set. Fortunately, most regression tree libraries use some techniques, as cost complexity pruning \cite{gareth:2021}, for controlling the degree of partitioning and therefore achieving a good bias/variance balance.

\section{Regression trees for univariate time series forecasting}
\label{sec_rt_utsf}

This section describes a general approach for univariate time series forecasting using regression trees. In an univariate setting only the historical values of the series are used to forecast its next future values, i.e., there are no exogenous variables \cite{fpp3}. The usual way to proceed is to build an autoregressive model in which the response variable (or target) is explained by means of previous values of the variable in the series. These previous or lagged values of a variable are also called \textit{lags}. For example, given the time series $t = \{t_1, t_2, t_3, t_4, t_5, t_6, t_7, t_8\}$ and supposing that we want to build an autoregressive model in which a target is explained by its three previous values, Table \ref{tab_training_set} shows the training set associated with the series.

\begin{table}
	\centering
	\begin{tabular}{ |c|c|c|c|} 
		\hline
		Lag 3 & Lag 2 & Lag 1 & Target \\ \hline
		$t_1$ & $t_2$ & $t_3$ & $t_4$  \\ 
		$t_2$ & $t_3$ & $t_4$ & $t_5$  \\ 
		$t_3$ & $t_4$ & $t_5$ & $t_6$  \\ 
		$t_4$ & $t_5$ & $t_6$ & $t_7$  \\ 
		$t_5$ & $t_6$ & $t_7$ & $t_8$  \\ 
		\hline
	\end{tabular}
	\caption{Training set associated with series $\{t_1, t_2, t_3, t_4, t_5, t_6, t_7, t_8\}$ and autoregressive lags 1 to 3.}
	\label{tab_training_set}
\end{table}

Suppose the previous series, $t$, and an autoregressive model trained with its associated training set---the one shown in Table \ref{tab_training_set}. In the case that we want to predict the next future value of the series, its vector of features would be $(t_6, t_7, t_8)$. The model would be fed with this input vector to generate the forecast for the next future value of the series. When the forecasting horizon (the number of future values to be forecast) is greater than one, there are several ways to produce the predictions, most notably the direct and recursive strategies \cite{Taieb:2012}. In this work we will use the recursive strategy, because it is simple and effective \cite{Martinez:2019}. Using this approach, the one-step ahead value is forecast as explained previously, producing the value $f_1$. To generate the two-steps ahead prediction the input point would be $(t_7, t_8, f_1)$. Note that, since the previous value to the value being forecast is unknown, its predicted value ($f_1$) is used. Table \ref{tab_recursive_forecast} shows the input points and forecasts for the different $n$-steps ahead predictions for a forecasting horizon of 4 using the recursive approach and the previous example.

\begin{table}
	\centering
	\begin{tabular}{ |c|c|c|} 
		\hline
		Steps ahead forecast & Input point & Forecast \\ \hline
		1 & $(t_6, t_7, t_8)$ & $f_1$  \\ 
		2 & $(t_7, t_8, f_1)$ & $f_2$  \\ 
		3 & $(t_8, f_1, f_2)$ & $f_3$  \\ 
		4 & $(f_1, f_2, f_3)$ & $f_4$  \\ 
		\hline
	\end{tabular}
	\caption{Input points for series $\{t_1, t_2, t_3, t_4, t_5, t_6, t_7, t_8\}$, autoregressive lags 1 to 3, the recursive strategy and forecast horizon 4.}
	\label{tab_recursive_forecast}
\end{table}

Once explained how autoregressive models and the recursive prediction work, let us see how these tools are integrated in an univariate time series forecasting example using a regression tree. We will use the artificial quarterly series in Fig. \ref{fig_seasonal_series}---the black line. This series contains 20 observations corresponding to five whole years, from 2019 to 2023. We can observe that:

\begin{figure}
	\centering
	\includegraphics[scale=0.6]{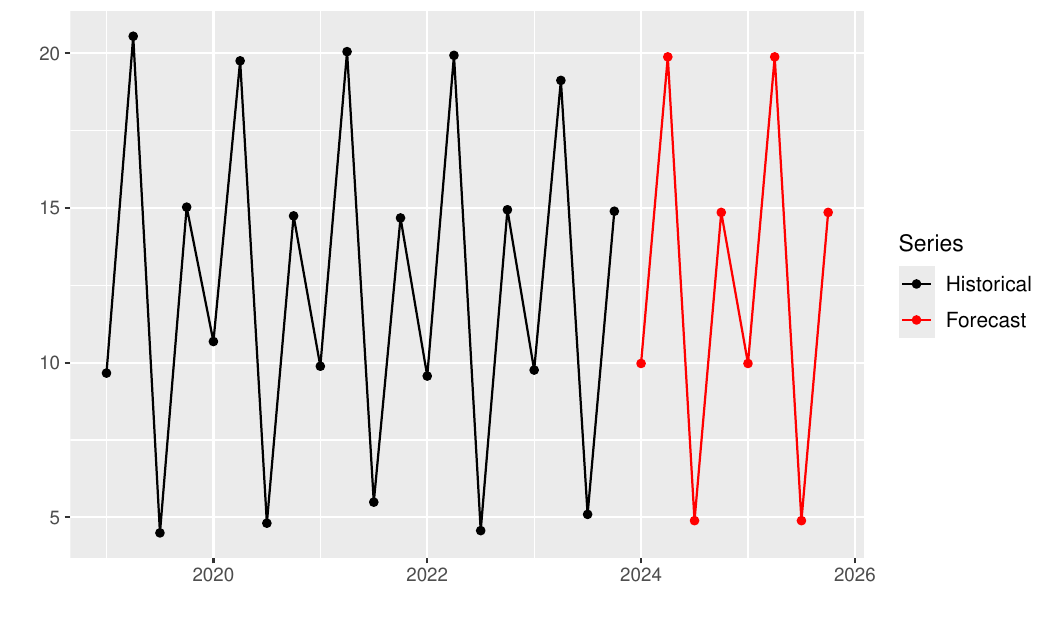}
	\caption{An artificial quarterly time series and its forecast.}
	\label{fig_seasonal_series}
\end{figure}

\begin{itemize}
	\item The level of the different values in every quarter is similar. The level of quarter 1 is about 10, quarter 2 about 20, quarter 3 about 5 and quarter 4 about 15.
	\item The ranges of the different quarters do not overlap.
\end{itemize}

\begin{figure}
	\centering
	\includegraphics[scale=0.5]{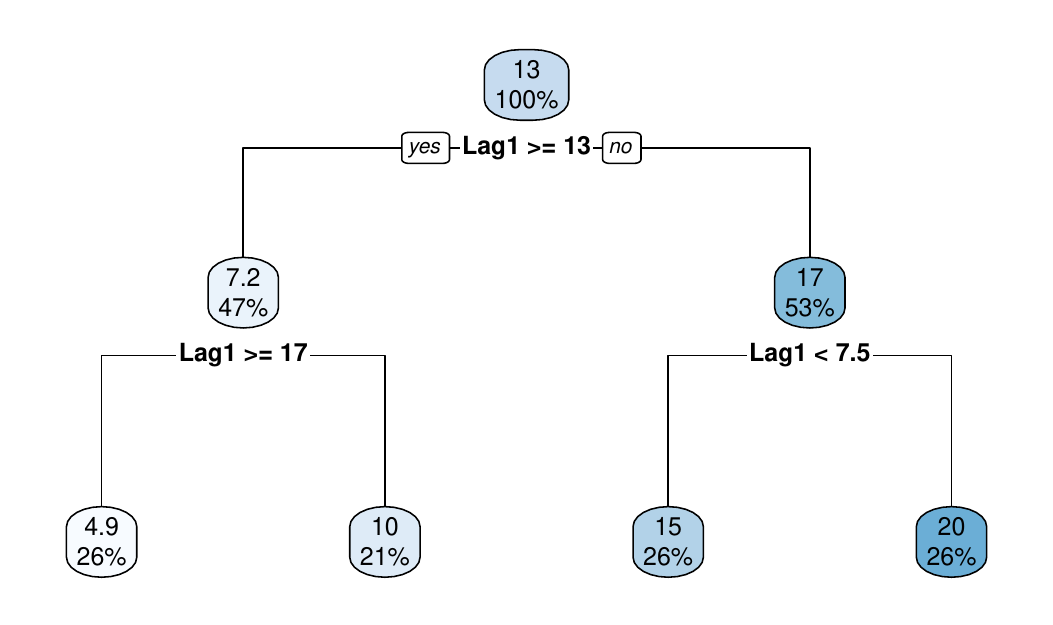}
	\caption{An autoregressive tree for the time series in Fig. \ref{fig_seasonal_series} based on the previous value of the response.}
	\label{fig_rt_model_seasonal1}
\end{figure}

This two observations suggest that an effective autoregressive tree model can be built wherein a future value is explained by its previous value. This model is shown in Fig. \ref{fig_rt_model_seasonal1} and its forecast for the next two years appears in red in Fig. \ref{fig_seasonal_series}. The tree has four leaves and, therefore, can only predict four different values. The forecast perfectly captures the seasonal pattern of the series.

The regression tree can be represented by four rules:
\begin{enumerate}
	\item if $Lag1 < 13$ and $Lag1 < 7.5$, predict 15.
	\item if $Lag1 < 13$ and $Lag1 \geq 7.5$, predict 20.
	\item if $Lag1 \geq 13$ and $Lag1 \geq 17$, predict 4.9.
	\item if $Lag1 \geq 13$ and $Lag1 < 17$, predict 10.
\end{enumerate}

\begin{figure}
	\centering
	\includegraphics[scale=0.5]{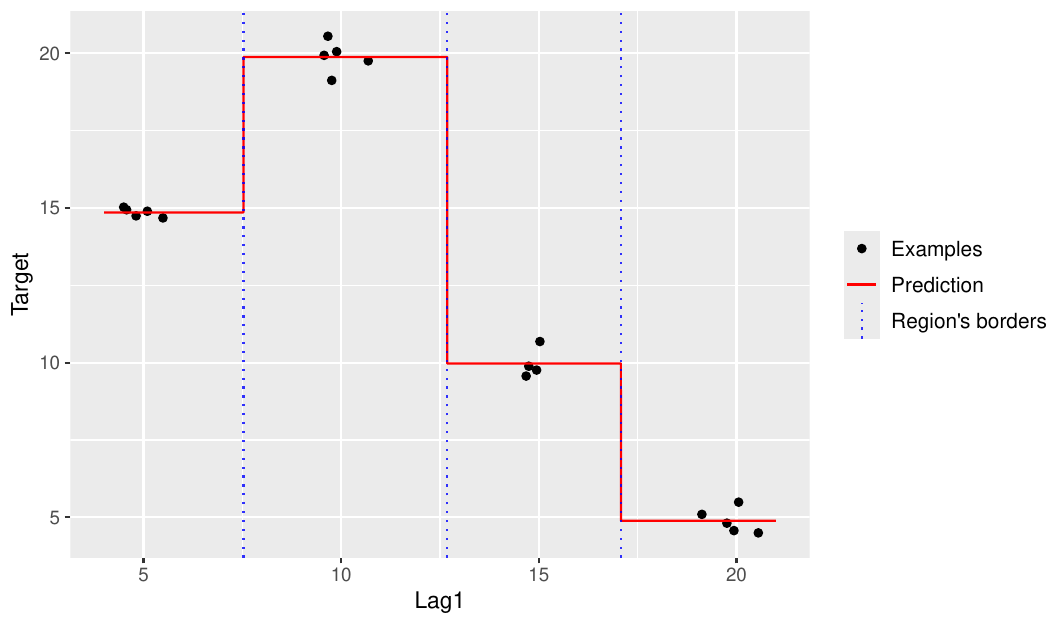}
	\caption{Feature space partition and predictions associated with regression tree in Fig. \ref{fig_rt_model_seasonal1}.}
	\label{fig_seasonal_fsp}
\end{figure}

Fig. \ref{fig_seasonal_fsp} shows the feature space partition associated with the regression tree, the red line represents the average value of the targets in the region.

\section{Bagging and random forests}
\label{sec_bagging_rf}

Bagging \cite{Breiman:1996} and random forests \cite{Breiman:2001} are two well-known ensemble approaches applied to decision trees to improve their robustness and prediction accuracy. Both approaches build a set of diverse deep decision trees and average the predictions of the individual trees to generate a smoothed final forecast that reduces the variance associated with regression trees. Bagging introduces diversity among the trees using bootstrapping, so that each tree is trained with a different bootstrapped sample of the training set. Unfortunately, if there are dominant predictors, the trees of a bagged ensemble are highly correlated, limiting the effectiveness of the aggregation process in reducing variance. Random forests also uses bootstrapping to select the training sets of the decision trees; furthermore, they try to avoid the potential high correlation among the trees adding more randomness in the creation of the trees: each time a split is to be performed, a random subset of  the original features is used as candidate set.

Bagging and random forests are interesting models for an automatic tool because, apart from improving forecast accuracy, they tend to offer a good performance with a default configuration, i.e., without hyperparameter tuning.

\section{Dealing with trending series}
\label{sec_trend}

\begin{figure}
	\centering
	\includegraphics[scale=0.5]{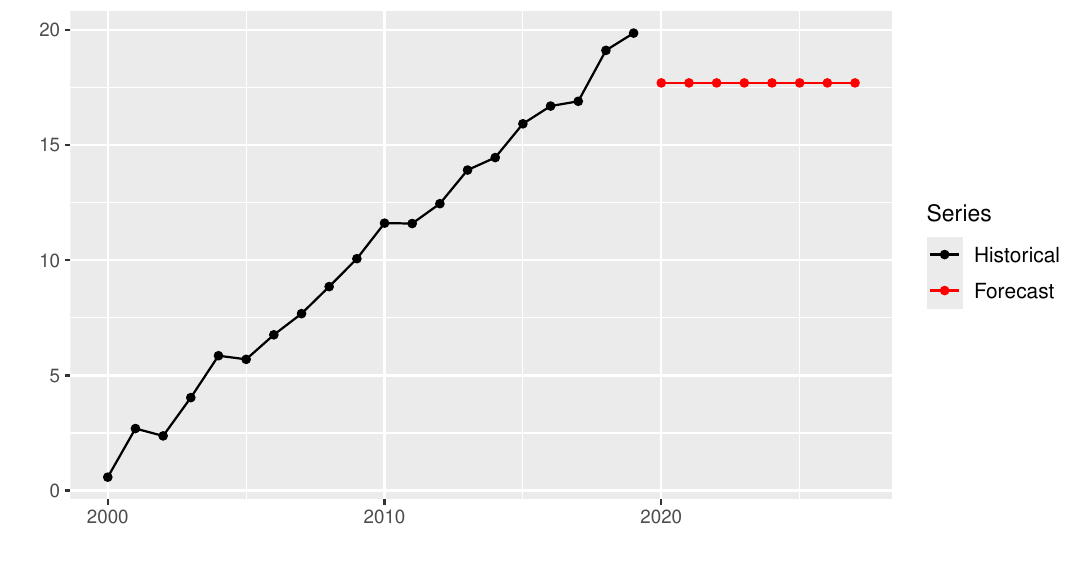}
	\caption{A trending series and its forecast using a regression tree.}
	\label{fig_trending}
\end{figure}

\begin{figure}
	\centering
	\includegraphics[scale=0.45]{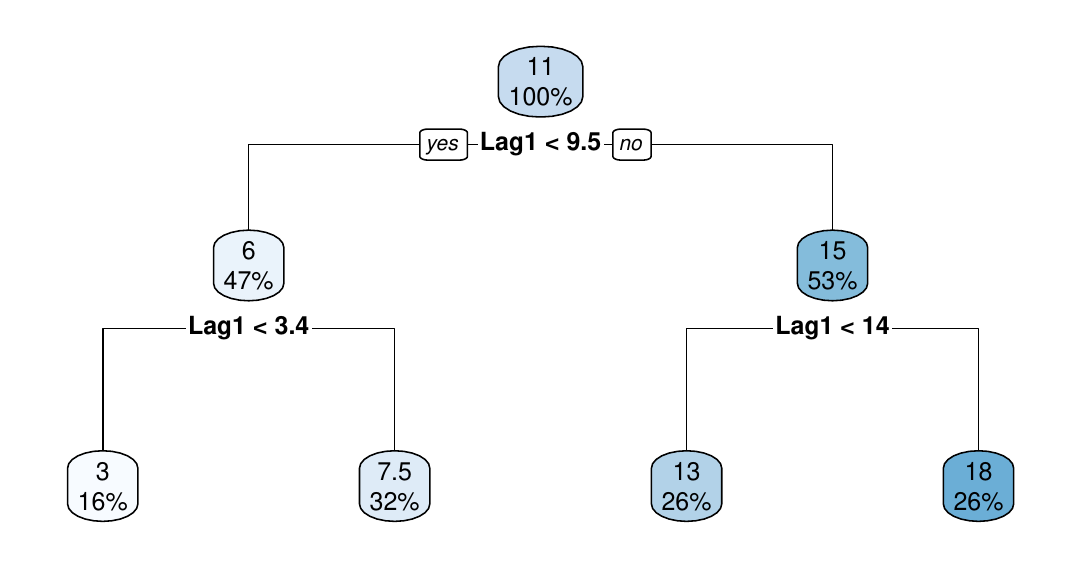}
	\caption{Regression tree for series in Fig. \ref{fig_trending} using lag 1 as autoregressive feature.}
	\label{fig_rt_trending}
\end{figure}

\begin{figure}
	\centering
	\includegraphics[scale=0.45]{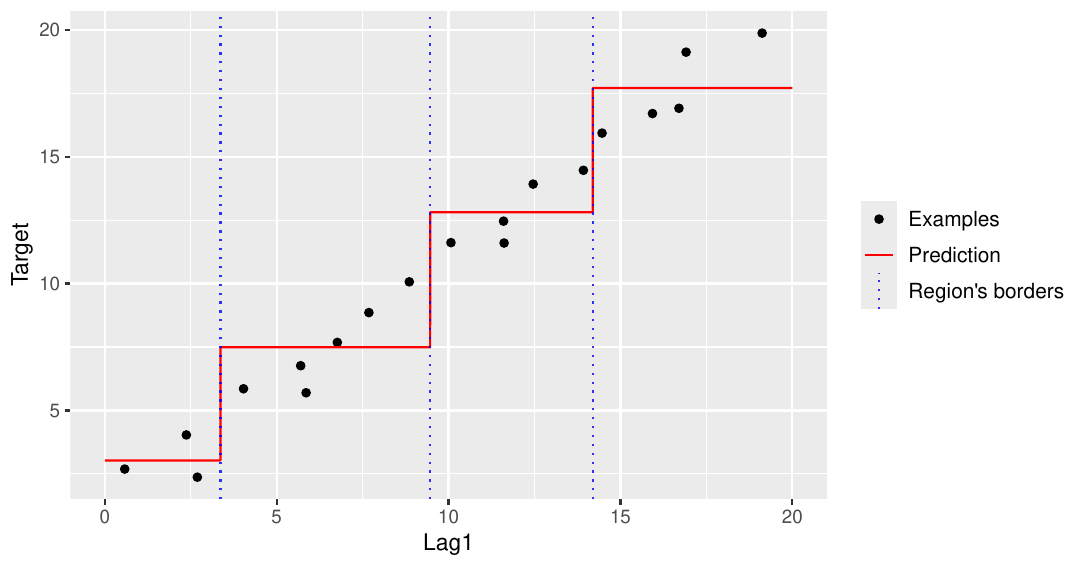}
	\caption{Feature space partition and prediction for regression tree in Fig. \ref{fig_rt_trending}.}
	\label{fig_fsp_trending}
\end{figure}

Regression trees predict averaged values of the target variable. Therefore, in an autoregressive context for time series forecasting, they can only generate forecasts in the range of the historical values of a series. Let us see an example, Fig. \ref{fig_trending} shows an artificial series with an upward trend and its forecast for horizon eight using the regression tree in Fig. \ref{fig_rt_trending}, which uses lag 1 as autoregressive feature. Its associated feature space subdivision and predictions for every region of the subdivision are shown in Fig. \ref{fig_fsp_trending}. When the regression tree is recursively used to forecast the next eight future values, the input features fall in the rightmost region in the space partition in Fig. \ref{fig_fsp_trending} and the value 18 is always produced as prediction. 

Clearly, some kind of strategy is needed in order to properly forecast trending series with regression trees. This work explores three approaches: differencing and two transformations of the training examples. Let us see these strategies:

\subsection*{Differencing}

It is possible to remove the trend in a series by taking first differences, that is, subtracting from each value of the series its previous value \cite{ord2017principles}. Then, a regression tree can be trained with the differenced series and the forecast generated can be back-transformed to add the trend to the prediction. Sometimes, some trend remains in a first differenced series and a second level of differencing is needed. There are statistical tests to estimate the number of first differences required to make a time series stationary and, thus, remove its trend.

\subsection*{Additive transformation}

This transformation has been applied successfully in previous works to forecast trending series using generalized regression neural networks and k-nearest neighbors \cite{Martinez:2022, Martinez:2025}. It consists of:

\begin{itemize}
	\item Transform every target of the training examples, subtracting from it the mean of its associated vector of features. Thus, a transformed target represents the difference between the original target and the mean level of its associated vector of features.
	\item A forecast is back-transformed by adding to it the mean of the input vector.
	\item Optionally, when the number of autoregressive lags is greater than one, the vector of features of every example can be transformed by subtracting the mean of the vector. The goal of this transformation is to remove the effect of the level of the series in the examples. In this way, the similarity between features in the feature space in a trending series is not dominated by the level of their associated examples.
\end{itemize}	

\subsection*{Multiplicative transformation}

It is similar to the additive transformation, but aimed at series with an exponential trend, rather than a linear trend. It consists of:

\begin{itemize}
	\item Every target of the training examples is divided by the mean of its associated vector of features.
	\item A forecast is back-transformed by multiplying it by the mean of the input vector.
	\item Optionally, when the number of autoregressive lags is greater than one, the vector of features of every example can be transformed by dividing it by the mean of the vector. 
\end{itemize}

\begin{figure}
	\centering
	\includegraphics[scale=0.45]{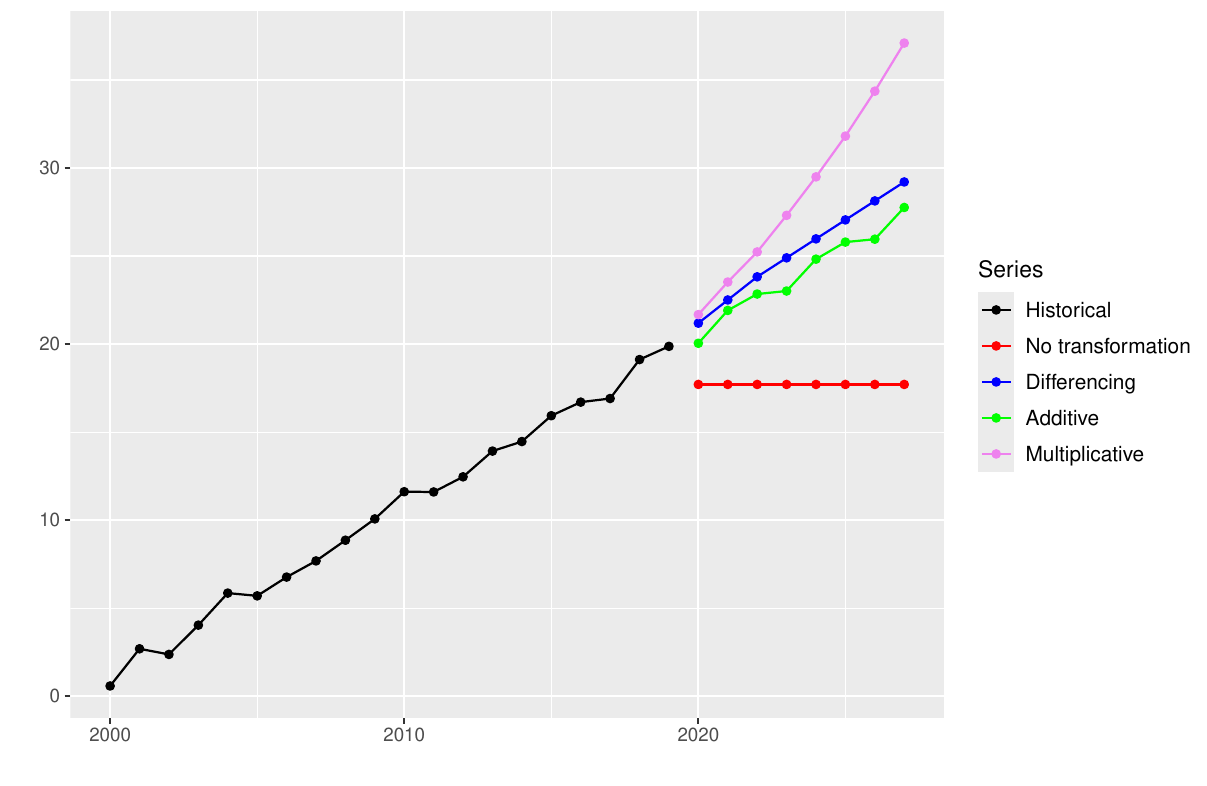}
	\caption{Forecasting a trending series with different strategies.}
	\label{fig_trend_comparison}
\end{figure}

Fig. \ref{fig_trend_comparison} shows how the series in Fig. \ref{fig_trending} is predicted using regression trees that apply the three strategies for dealing with trending series recently explained. Both the forecasts taking first differences and using the additive transformation seem to capture the linear trend. As was previously seen, if no transformation is applied the forecast underestimates the upward trend. On the other hand, the forecast associated with the multiplicative transformation seems to be over-optimistic for a linear trend.
 
In order to compare these strategies for dealing with trending series an experimentation has been done, in which the 23,000 yearly series from the M4 competition \cite{MAKRIDAKIS:2020} are forecast. This dataset is particularly interesting, because many yearly series have a trend. We have downloaded the dataset from the Monash Time Series Forecasting Repository \cite{godahewa:2021}, this repository not only provides datasets, but a framework in which your results can be compared with other time series forecasting models. In order to obtain results comparable to those of the repository, we have computed forecast accuracy using the MASE: the Mean Absolute Scaled Error \cite{HYNDMAN:2006}. MASE is scale independent and, therefore, can be used to compare the forecast accuracy across different series. MASE computes the mean absolute error of a forecast relative to the mean absolute error of a simple forecasting method for one-step ahead forecasts on the training data (the historical values of the series). For example, given a series $\{h_1, h_2, \ldots, h_T\}$ and a forecast $\{\hat{y}_1, \hat{y}_2, \ldots, \hat{y}_h\}$ for its next $h$ future values, the accuracy of the forecast is computed as follows:

$$
\textit{MASE} = \frac{\frac{1}{h}\sum_{t=1}^{h} |y_t - \hat{y}_t|}{\frac{1}{T-f}\sum_{t=f+1}^{T} |h_t - h_{t-f}|}
$$

where $y_n$ is the actual $n$-steps ahead future value and $f$ is the frequency of the series. For yearly series $f$ is 1, so MASE indicates the mean absolute error of the forecast for horizon $h$, relative to the mean absolute one-step ahead error on the training set (historical values) for the naive method, i.e.,  when the forecasting method is to predict the next future value as its previous value in the series.

For quarterly series $f$ is 4 and, therefore, the simple forecast for a quarter is its previous known value for that quarter. For monthly series $f$ is 12.

The M4 yearly dataset contains 23,000 series with lengths from 13 to 835. The forecast horizon for all the series is 6. To obtain an estimation of the forecast accuracy of a method on the dataset, every series of the dataset is forecast and its MASE computed and, then, the mean and median MASE values across all the series in the dataset is calculated. These values are shown in Table \ref{tab_trend_strategies_comparison}. All the forecasting models are regression trees with lags 1 to 5 as autoregressive features. The first column indicates the strategy used for dealing with trends. Let us make some comments:

\begin{itemize}
	\item The worst results are clearly obtained when no transformation is applied. This is expected, because in about 76\% of the series in the M4 yearly dataset, the range of the test set is outside the range of the historical values. The fact that applying no transformation yields the worst results suggests that the strategies for dealing with trends are effective.
	\item The additive transformation obtains the best results. Furthermore, it seems that the transformation of the features (subtracting its values from their mean value) has a positive impact.
	\item The multiplicative transformation seems to be less effective than the other strategies.
	\item Compared to other strategies, differencing obtains better results in median MASE than in mean MASE, indicating bad results for some series. Unlike the additive and multiplicative transformation, that are applied to all the series, first differences are only applied to those series in which a statistical test estimates that the series are not stationary. Concretely, the \textit{nndiffs} function from the \textit{forecast} R package  \cite{Hyndman:2008} is used to estimate the number of first differences required.
\end{itemize}

\begin{table}
	\centering
	\begin{tabular}{ |l|c|c|} 
		\hline
		\textbf{Strategy} & \textbf{Mean MASE} & \textbf{Median MASE} \\ \hline
		No transformation & 7.902 & 6.579 \\
		Additive transformation of features \& targets & \textbf{3.387} & \textbf{2.468} \\
		Additive transformation of targets & 3.528 & 2.551 \\
		Multiplicative transformation of features \& targets & 4.563 & 3.301 \\
		Multiplicative transformation of targets & 4.651 & 3.286 \\
		Differencing & 4.020 & 2.667 \\
		\hline
	\end{tabular}
	\caption{Mean and median MASE for the M4 yearly dataset for a regression tree with different strategies for dealing with trending series.}
	\label{tab_trend_strategies_comparison}
\end{table}

In general, the results achieved by the regression tree using the additive transformation applied to both features and targets are quite competitive. For example, it is better than all the methods evaluated in \cite{godahewa:2021}. As an additional experimentation, we have used bagging and random forests with regression trees using the additive transformation (for both features and targets) and we have also applied two classical statistical univariate time series forecasting models as exponential smoothing and ARIMA for benchmarking purposes. The results are shown in Table \ref{tab_trend_others_comparison}. We would like to make some comments:

\begin{itemize}
	\item Both bagging and random forests are stochastic methods, so their results are not deterministic.
	\item Bagging only slightly improves the result of the base learner (the regression tree with the additive transformation). The ensemble uses 25 trees.
	\item Random forests achieves the best results among all the models. It uses 500 trees and randomly considers one third of the features in each split.
	\item Exponential smoothing has been applied using the \textit{ets} function from the \textit{forecast} R package \cite{Hyndman:2008}. This function automatically selects the best exponential smoothing model among nine candidates. 
	\item ARIMA has been applied using the \textit{auto.arima} function from the \textit{forecast} R package. This function automatically selects the best ARIMA model using optimization. 
\end{itemize}

\begin{table}
	\centering
	\begin{tabular}{ |l|c|c|} 
		\hline
		\textbf{Strategy} & \textbf{Mean MASE} & \textbf{Median MASE} \\ \hline
		Bagging & 3.346 & 2.409 \\
		Random forests & \textbf{3.170} & \textbf{2.277} \\
		Exponential smoothing & 3.444 & 2.329 \\
		ARIMA & 3.401 & 2.314 \\
		\hline
	\end{tabular}
	\caption{Mean and median MASE for the M4 yearly dataset for different models.}
	\label{tab_trend_others_comparison}
\end{table}

\section{Dealing with seasonality}
\label{sec_seasonality}

Time series can be affected by periodic time factors, such as the day of the week or the hour of the day. A forecasting model has to capture these seasonal patterns in order to effectively predict a series exhibiting \textit{seasonal behavior}.

\begin{figure}
	\centering
	\includegraphics[scale=0.45]{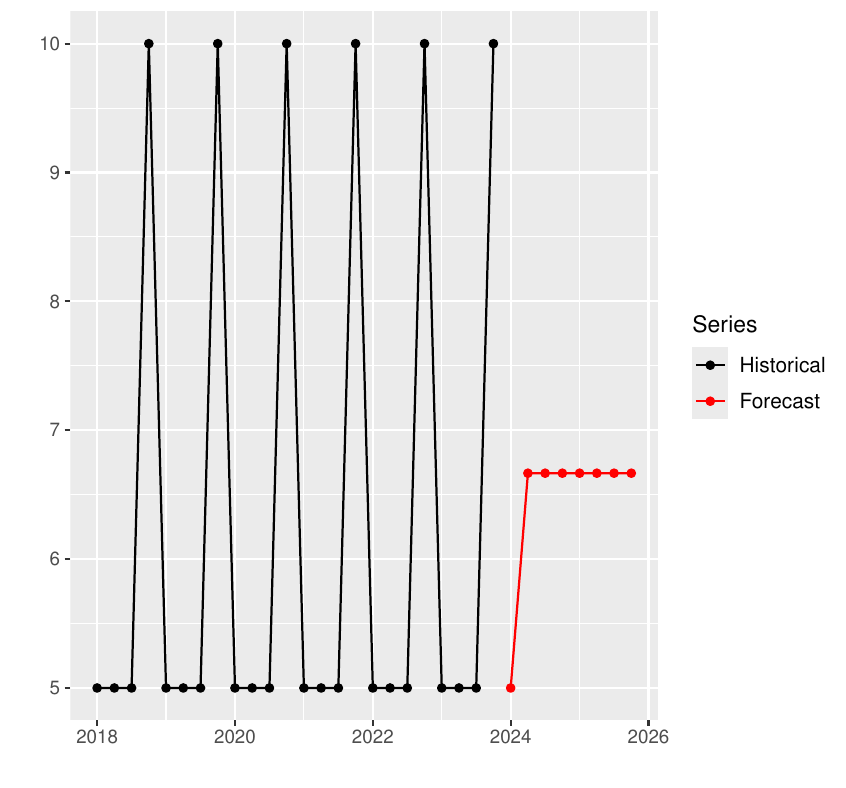}
	\caption{Artificial quarterly series and its forecast using a regression tree with lag 1 as autoregressive feature.}
	\label{fig_qs_forecast_lag1}
\end{figure}

\begin{figure}
	\centering
	\includegraphics[scale=0.45]{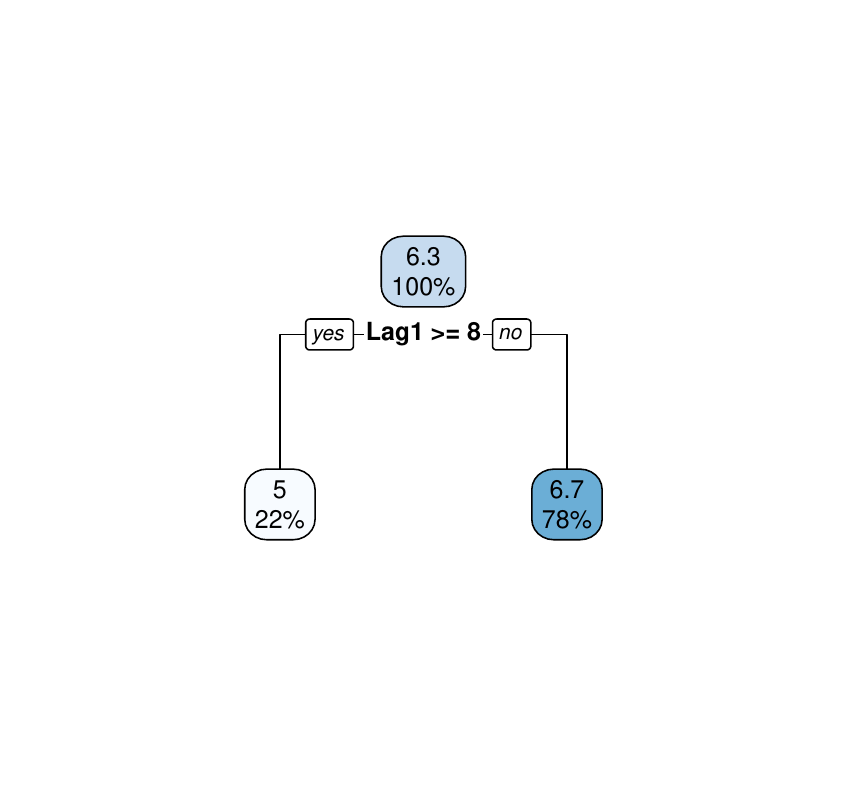}
	\caption{Regression tree associated with the forecast in Fig. \ref{fig_qs_forecast_lag1}.}
	\label{fig_qs_model_lag1}
\end{figure}

\begin{figure}
	\centering
	\includegraphics[scale=0.45]{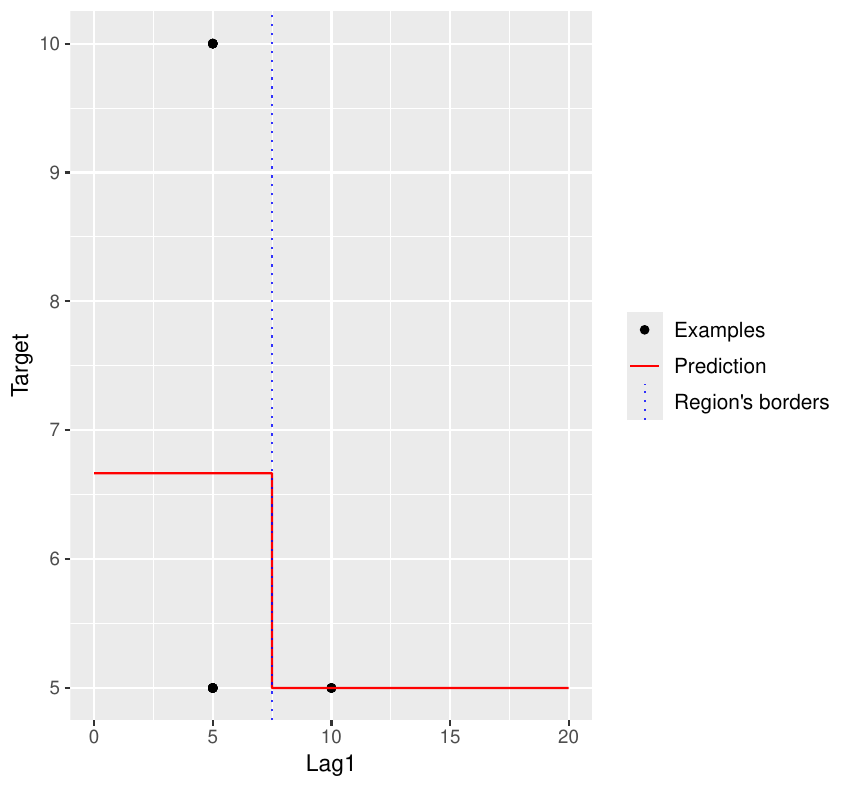}
	\caption{Feature space subdivision and predictions associated with the regression tree in Fig. \ref{fig_qs_model_lag1}.}
	\label{fig_qs_feature_space_lag1}
\end{figure}

\begin{figure}
	\centering
	\includegraphics[scale=0.45]{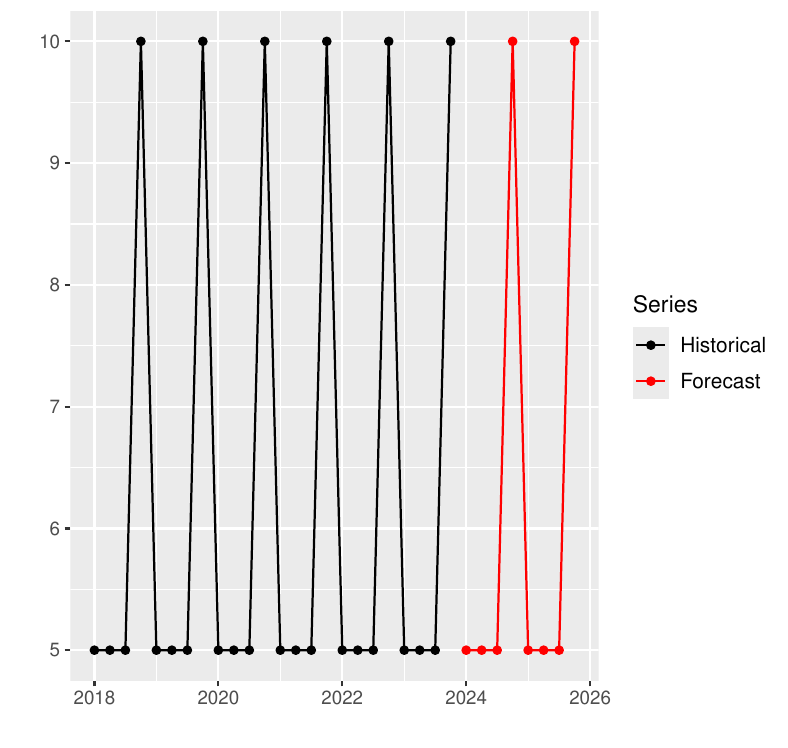}
	\caption{Artificial quarterly series and its forecast using a regression tree with lag 4 as autoregressive feature.}
	\label{fig_qs_forecast_lag4}
\end{figure}

\begin{figure}
	\centering
	\includegraphics[scale=0.45]{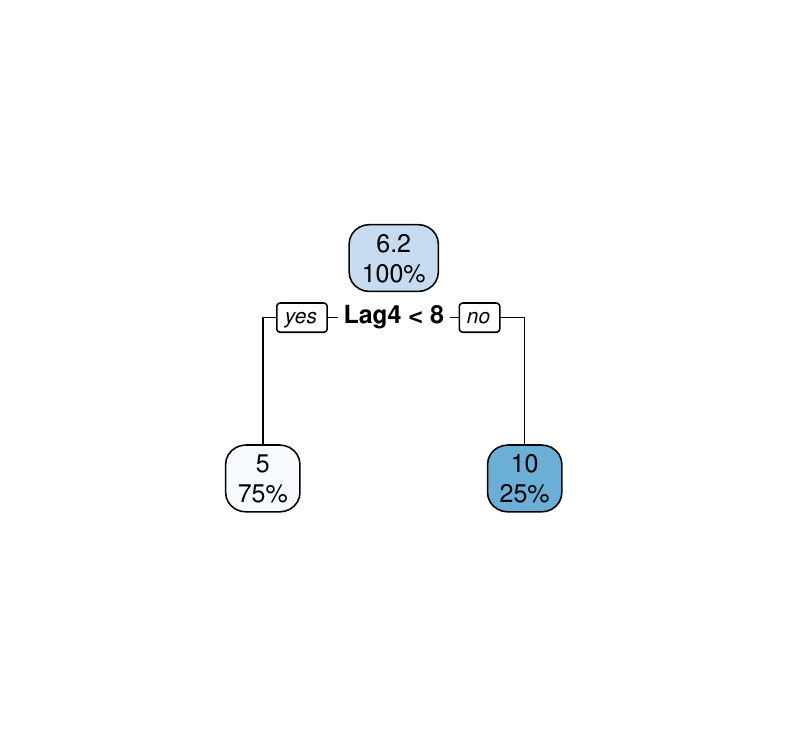}
	\caption{Regression tree associated with the forecast in Fig. \ref{fig_qs_forecast_lag4}.}
	\label{fig_qs_model_lag4}
\end{figure}

In the case of our autoregressive models it is important that the autoregressive lags include the length of the seasonal period. For example, a monthly series should include lag 12 and a daily series with weekly seasonality should include lag 7. Let us see an extreme example using regression trees that illustrates why it is crucial to use the length of the seasonal period as autoregressive lag. Fig. \ref{fig_qs_forecast_lag1} shows an artificial quarterly series in which all the first three quarters of a year have the same value (5) and the fourth quarter has value 10. The figure also shows the forecast of a regression tree using lag 1 as autoregressive feature. Clearly, the model does not capture the seasonal series behavior. This model is shown in Fig. \ref{fig_qs_model_lag1}. It is quite simple because the features of the training set only have two possible values (5 and 10). The problem is that lag 1 is not a good predictor because for the fourth, third and second quarters the lagged value is the same: 5. However, the fourth quarter has a value of 10 and the third and second quarter a different value: 5. Therefore, lag 1 cannot predict correctly the next future value of the series. Fig. \ref{fig_qs_feature_space_lag1} shows the feature space partition associated with the regression tree in Fig. \ref{fig_qs_model_lag1} and the predictions for every region of the partition. The left region is the problematic one because it contains quite dissimilar targets. If lag 2 or lag 3 were used instead of lag 1 as autorregresive features the same problem arises. However, as a consequence of the strong seasonal behavior of the series, lag 4 is a good predictor, as can be seen in Fig. \ref{fig_qs_forecast_lag4} in which the forecast for the example artificial quarterly series using a regression tree with lag 4 as feature is shown. The regression tree associated with the model is shown in Fig. \ref{fig_qs_model_lag4}.

We have done some experimentation to see how the forecast accuracy is affected by the lags used as autoregressive variables. As dataset we have used the 24,000 quarterly series from the M4 competition \cite{MAKRIDAKIS:2020}, because quarterly series are prone to exhibit seasonal patterns. The lengths of the series range from 16 to 866. For all the series the forecasting horizon is 8. Since the series are quarterly, we have used lags one to four as possible autoregressive variables. The additive transformation has been applied to both the features and the targets. The result of this comparison for regression tree models is shown in Table \ref{tab_seasonal_comparison}. As can be noted, the addition of lags improves the accuracy of the models, with the best results achieved by the model with autoregressive lags 1 to 4. 

In this case, we have also experimented with using bagging and random forest to check whether they can improve regression trees. For both models we have used autoregressive lags from 1 to 4 and the additive transformation for both features and targets. The results are shown in Table \ref{tab_seasonal_others_comparison}. As can be seen, the use of both models seem to improve the forecast accuracy. Table \ref{tab_seasonal_others_comparison} also shows the forecast accuracy of the exponential smoothing and ARIMA implementations described previously. For the quarterly data the statistical models obtain the best results.

\begin{table}
	\centering
	\begin{tabular}{|c|c|c|} 
		\hline
		\textbf{Autoregressive lags} & \textbf{Mean MASE} & \textbf{Median MASE} \\ \hline
		1 & 1.582 & 1.165 \\
		1, 2 & 1.470 & 1.110 \\
		1, 2, 3 & 1.397 & 1.079 \\
		1, 2, 3, 4 & \textbf{1.344} & \textbf{1.032} \\
		\hline
	\end{tabular}
	\caption{Mean and median MASE for the M4 quartely dataset for a regression tree with different autoregressive lags.}
	\label{tab_seasonal_comparison}
\end{table}

\begin{table}
	\centering
	\begin{tabular}{ |l|c|c|} 
		\hline
		\textbf{Strategy} & \textbf{Mean MASE} & \textbf{Median MASE} \\ \hline
		Bagging & 1.248 & 0.957 \\
		Random forests & 1.203 & 0.929 \\
		Exponential smoothing & \textbf{1.161} & \textbf{0.886} \\
		ARIMA & 1.165 & 0.895 \\
		\hline
	\end{tabular}
	\caption{Mean and median MASE for the M4 quarterly dataset for different models.}
	\label{tab_seasonal_others_comparison}
\end{table}

\section{Our proposal}
\label{sec_proposal}

This section describes our proposal for automatic forecasting of univariate time series using regression trees. We suggest the following strategies:

\begin{itemize}
	\item Apply the additive transformation for both  features and targets. This transformation is useful for series with level changes.
	\item Selection of autoregressive lags. If the series is seasonal, use lags from 1 to the length of the seasonal period. For example, 1 to 4 for quarterly series or 1 to 12 for monthly series. If the series is not seasonal, use the significant autocorrelated lags in the partial autocorrelation function. If no lag has a significant autocorrelation, then lags from 1 to 5 can be chosen. An interesting property of regression trees is that they only use the relevant features of a training set (a kind of automatic feature selection), so if you doubt, you can try a generous amount of lags.
	\item Use random forests to reduce the variance and obtain more robust and accurate forecasts. Random forests work quite well with default parameters so they can be easily integrated in an automatic tool. Be aware, however, that random forests demand more computational time.
\end{itemize}

\section{Software: the utsf package}
\label{sec_software}

This section briefly describes the \texttt{utsf} R package, which implements the different strategies proposed in this paper. The package can be downloaded from CRAN, the official repository of R packages:

\begin{verbatim}
> install.packages("utsf")
\end{verbatim}

The \texttt{utsf} package is an engine that allows you to forecast time series applying different classical regression models using the autoregressive approach and the recursive forecasts explained in Section \ref{sec_rt_utsf}. All the regression models are applied through a common interface consisting of  two functions: \texttt{create\_model} and \texttt{forecast}. The \texttt{create\_model} function takes a time series, some autoregressive lags and a type of regression model and builds a training set from the series using the autoregressive lags. Then, it fits a model with the training set. For example:

\begin{verbatim}
> library(utsf)
> t <- ts(1:10)
> m <- create_model(t, lags = 1:3, method = "rt", trend = "none")
\end{verbatim}

In this case, the autoregressive lags 1 to 3 have been applied to the series \texttt{t} to create the training set. It is possible to see the features and targets of the training set:

\begin{verbatim}
> cbind(m$features, Target = m$targets)
  Lag3 Lag2 Lag1 Target
1    1    2    3      4
2    2    3    4      5
3    3    4    5      6
4    4    5    6      7
5    5    6    7      8
6    6    7    8      9
7    7    8    9     10
\end{verbatim}

The fitted model is a regression tree (\texttt{method = "rt"}) which is stored in the component named \texttt{model} of the object returned by the \texttt{create\_model} function:

\begin{verbatim}
> m$model
n= 7 

node), split, n, deviance, yval
* denotes terminal node
	
1) root 7 28 7 *
\end{verbatim}

In this case, the regression tree is a stump (a single-level tree) predicting the average value of all the targets (7). Because no pre-processing has been applied to deal with the trend (\texttt{trend = "none"}) the forecast will not capture the trending behavior. Once the model has been fitted, you can forecast future values of the series (using the recursive strategy) with the \texttt{forecast} function:

\begin{verbatim}
> f <- forecast(m, h = 4)
> f
Time Series:
Start = 11 
End = 14 
Frequency = 1 
[1] 7 7 7 7
\end{verbatim}

In this case, the forecast horizon is 4 (\texttt{h = 4}). It is possible to obtain a plot with the series and its forecast (see Fig. \ref{fig_example_autoplot}):

\begin{verbatim}
> library(ggplot2)
> autoplot(f)
\end{verbatim}

\begin{figure}
	\centering
	\includegraphics[scale=0.45]{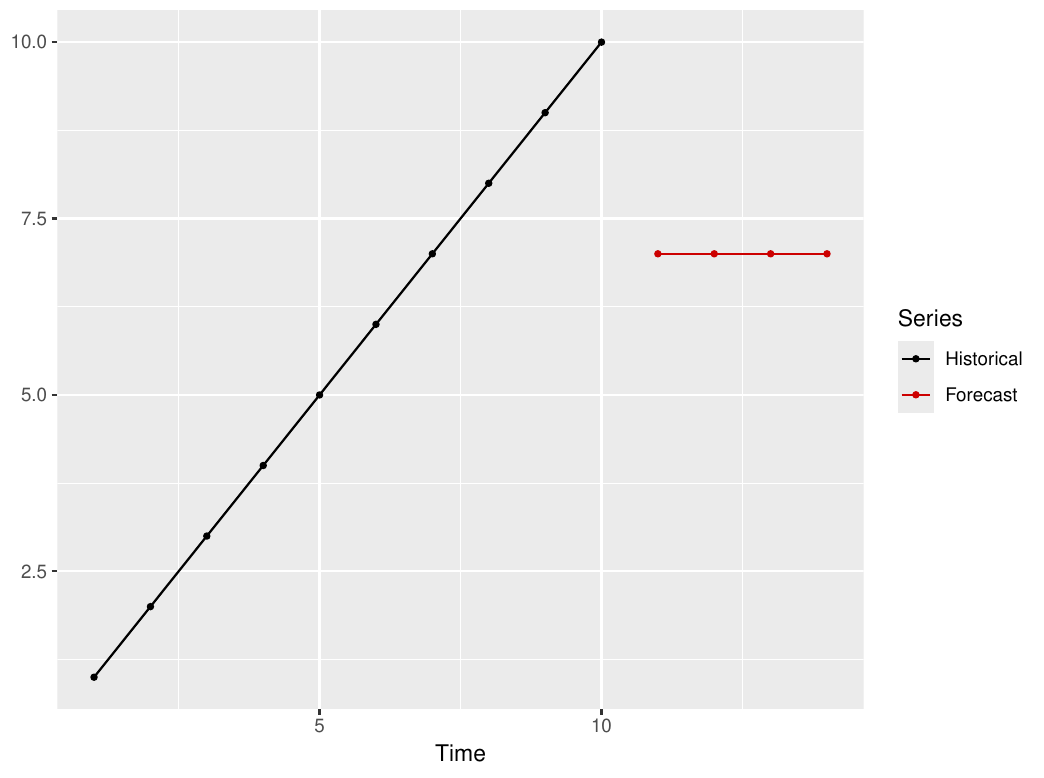}
	\caption{Time series and its forecast.}
	\label{fig_example_autoplot}
\end{figure}

\subsection{Supported models}

Currently, the \texttt{utsf} package supports the following regression models: regression trees, bagging, random forest, model trees, k-nearest neighbors and linear models. Model trees \cite{Q1992} is an interesting variant of regression trees in which a linear model is fit to the examples of every region of the feature space partition. We have used implementations of regression models in CRAN packages, for example, the \texttt{rpart} package is used for regression trees.

As mentioned before, the \texttt{utsf} package is an engine that allows you to use different regression models through a common interface. For example, next we fit three different models (regression tree, bagging and random forest):

\begin{verbatim}
> m1 <- create_model(t, lags = 1:3, method = "rt")
> m2 <- create_model(t, lags = 1:3, method = "bagging")
> m3 <- create_model(t, lags = 1:3, method = "rf")
\end{verbatim}

\subsection{Setting the parameters of the regression models}

Regression models can be tuned using different parameters. Because the \texttt{utsf} package is oriented toward automatic prediction, suitable parameters are selected by default. However, the user can choose specific parameters. For example:

\begin{verbatim}
> t <- ts(1:10)
> m <- create_model(t, lags = 1:3, method = "rt", trend = "none",
              param = list(control = rpart.control(minsplit = 3)))
> m$model              
n= 7 

node), split, n, deviance, yval
* denotes terminal node

1) root 7 28.0 7.0  
  2) Lag3< 3.5 3  2.0 5.0  
    4) Lag3< 1.5 1  0.0 4.0 *
    5) Lag3>=1.5 2  0.5 5.5 *
  3) Lag3>=3.5 4  5.0 8.5  
    6) Lag3< 5.5 2  0.5 7.5 *
    7) Lag3>=5.5 2  0.5 9.5 *
\end{verbatim}

In this case, the regression tree is built so that the minimum number of examples in a node for a split to be attempted is 3 (the default value is 20). In order to tune a regression model you must know the interface of the regression model used by the \texttt{utsf} package.

\subsection{Preprocessing for dealing with trends}

It is possible to apply the three strategies explained in Section \ref{sec_trend} to forecast a trending series. For example, let us see how to apply the additive transformation:

\begin{verbatim}
> t <- ts(1:10)
> m <- create_model(t, lags = 1:3, method = "rt", trend = "additive")
> cbind(m$features, Target = m$targets)
  Lag3 Lag2 Lag1 Target
1   -1    0    1      2
2   -1    0    1      2
3   -1    0    1      2
4   -1    0    1      2
5   -1    0    1      2
6   -1    0    1      2
7   -1    0    1      2
\end{verbatim}

In this case both targets and features have been transformed, as can be seen in the training set. You can transform only the targets:

\begin{verbatim}
> t <- ts(1:10)
> m <- create_model(t, lags = 1:3, method = "rt", trend = "additive",
                    transform_features = FALSE)
> cbind(m$features, Target = m$targets)
  Lag3 Lag2 Lag1 Target
1    1    2    3      2
2    2    3    4      2
3    3    4    5      2
4    4    5    6      2
5    5    6    7      2
6    6    7    8      2
7    7    8    9      2
\end{verbatim}

Now, let us see an example in which the three transformations are applied to a trending series trained with a random forest model:

\begin{verbatim}
> m_n <- create_model(airmiles, method = "rf", trend = "none")
> m_a <- create_model(airmiles, method = "rf", trend = "additive")
> m_m <- create_model(airmiles, method = "rf", trend = "multiplicative")
> m_d <- create_model(airmiles, method = "rf", trend = "differences")
> f_n <- forecast(m_n, h = 8)$pred
> f_a <- forecast(m_a, h = 8)$pred
> f_m <- forecast(m_m, h = 8)$pred
> f_d <- forecast(m_d, h = 8)$pred
> library(vctsfr)
> plot_predictions(airmiles, predictions = list(none = f_n,
                                                 additive = f_a,
                                                 multiplicative = f_m,
                                                 differences = f_d)
)
\end{verbatim}

\begin{figure}
	\centering
	\includegraphics[scale=0.45]{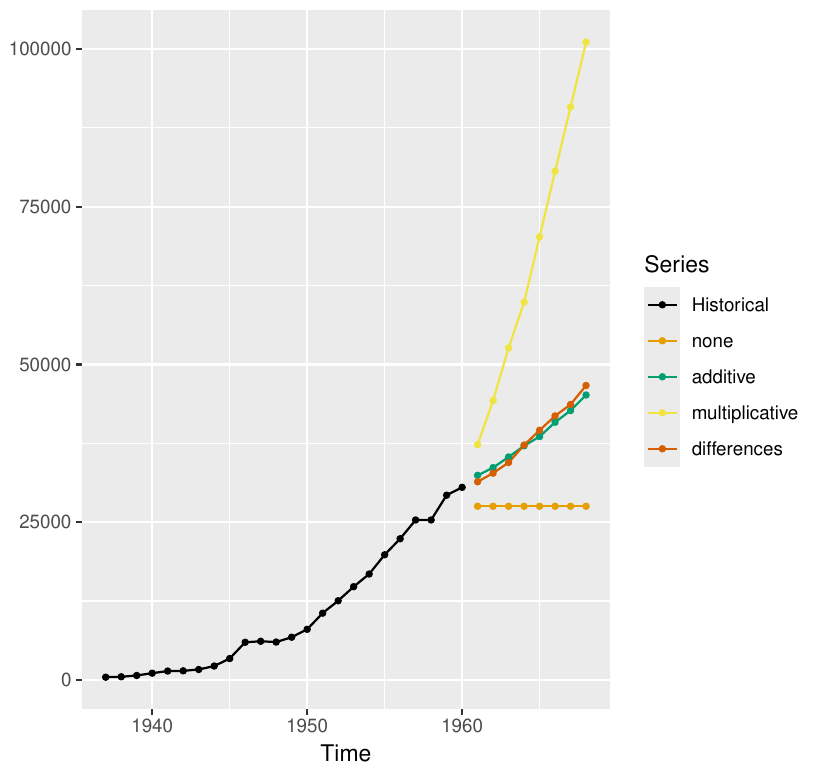}
	\caption{Time series and its forecasts using different strategies to cope with trends.}
	\label{fig_comparing_trend}
\end{figure}

Fig. \ref{fig_comparing_trend} shows the result of the different forecasts. Because the series has a trend, when no transformation is applied the trending pattern is not captured. On the other hand, the multiplicative transformation overestimates the trend. In this case, the additive transformation and taking differences produce similar, sensitive forecasts.

\subsection{Automated forecasts}

In order to forecast a series you only need to provide the chosen regression model, the autoregressive lags and the transformation is automatically selected as explained in Section
\ref{sec_proposal}. Also, sensible default parameters are selected for the regression models. For example, let us forecast a time series using a regression tree:

\begin{verbatim}
> create_model(USAccDeaths, method = "rt") |> forecast(h = 12)
          Jan      Feb      Mar      Apr      May      Jun      Jul
1979 8627.200 7217.481 8156.276 8750.696 9046.437 9040.723 9983.462
Aug      Sep      Oct      Nov      Dec
1979 9941.751 8975.801 8964.618 8706.654 8961.974
\end{verbatim}

and let us see the autoregressive lags and transformation applied:

\begin{verbatim}
> create_model(USAccDeaths, method = "rt")
Call:  create_model(timeS = USAccDeaths, method = "rt")

Autoregressive lags: 1 2 3 4 5 6 7 8 9 10 11 12 
Additive tranformation applied to features and targets
Regression model: regression trees
\end{verbatim}

\section{Conclusions}
\label{sec_conclusions}

This paper has described how classical regression tree models can be used to forecast univariate time series applying an autoregressive approach and recursive forecasts. In order to produce accurate predictions strategies to deal with trending and seasonal series are explained. Some of these strategies are suggested as a way to automatically predict time series. Specifically, the use of the additive transformation to both features and targets as a way to deal with trending series and, in general, with level shifts in a series. Also, how to select the autoregressive lags to capture potential seasonal behavior. Finally, the use of random forests is encouraged to reduce variance and improve forecast accuracy.

\bibliography{biblio}

@book{Breiman:1984,
   author = {Breiman, Leo and Friedman, J. H. and Olshen, R. A. and Stone, C. J.},
   title =  {Classification and regression trees},
   year = {1984},
   publisher =  {Chapman and Hall/CRC},
   url = {https://doi.org/10.1201/9781315139470}
}

@article{Breiman:1996,
	title = {{Baggin Predictors}},
	journal = {Machine Learning},
	volume = {24},
	number = {2},
	pages = {123-140},
	year = {1996},
	url = {https://doi.org/10.1007/BF00058655},
	author = {Breiman, Leo},
}

@article{Breiman:2001,
	title = {{Random Forests}},
	journal = {Machine Learning},
	volume = {45},
	number = {1},
	pages = {5-32},
	year = {2001},
	url = {https://doi.org/10.1023/A:1010933404324},
	author = {Breiman, Leo},
}

@article{Charte2020,
	author = {Francisco Charte Ojeda and Antonio Jesús Rivera Rivas and Francisco Martínez and Maria José del Jesus Díaz},
	title = {EvoAAA: An evolutionary methodology for automated neural autoencoder architecture search},
	year = {2020},
	journal = {Integrated Computer-Aided Engineering},
	volume = {27},
	number = {3},
	pages = {211-231},
	month = {05/2020},
	doi = {10.3233/ICA-200619},
}

@article{friedman2001,
	title={Greedy function approximation: a gradient boosting machine},
	author={Friedman, Jerome H},
	journal={The Annals of Statistics},
	volume={29},
	number={5},
	pages={1189--1232},
	year={2001},
	publisher={Institute of Mathematical Statistics},
	doi={10.1214/aos/1013203451}
}

@book{gareth:2021,
	author = {James, Gareth and Witten, Daniela and Hastie, Trevor and Tibshirani, Robert},
	title = {An Introduction to Statistical Learning: with Applications in R},
	year = {2021},
	publisher = {Springer Publishing Company, Incorporated},
	url={https://doi.org/10.1007/978-1-0716-1418-1},
	edition = {2nd}
}

@inproceedings{godahewa:2021,
	title={Monash Time Series Forecasting Archive},
	author={Rakshitha Wathsadini Godahewa and Christoph Bergmeir and Geoffrey I. Webb and Rob Hyndman and Pablo Montero-Manso},
	booktitle={Thirty-fifth Conference on Neural Information Processing Systems Datasets and Benchmarks Track (Round 2)},
	year={2021},
	url = {https://forecastingdata.org/}
}

@article{HUSSEIN2024,
	title = {Time series forecasting of electricity consumption using hybrid model of recurrent neural networks and genetic algorithms},
	journal = {Measurement: Energy},
	volume = {2},
	pages = {100004},
	year = {2024},
	issn = {2950-3450},
	doi = {https://doi.org/10.1016/j.meaene.2024.100004},
	url = {https://www.sciencedirect.com/science/article/pii/S2950345024000046},
	author = {Ali Hussein and Mohammed Awad},
}

@article{HYNDMAN:2006,
	title = {Another look at measures of forecast accuracy},
	journal = {International Journal of Forecasting},
	volume = {22},
	number = {4},
	pages = {679-688},
	year = {2006},
	url = {https://doi.org/10.1016/j.ijforecast.2006.03.001},
	author = {Rob J. Hyndman and Anne B. Koehler},
}

@Article{Hyndman:2008,
	title = {Automatic time series forecasting: the forecast package for {R}},
	author = {Rob J Hyndman and Yeasmin Khandakar},
	journal = {Journal of Statistical Software},
	volume = {27},
	number = {3},
	pages = {1--22},
	year = {2008},
	url = {https://doi.org/10.18637/jss.v027.i03},
}

@book{fpp3,
	title = "Forecasting: Principles and Practice",
	author = "Rob Hyndman and G. Athanasopoulos",
	year = "2021",
	language = "English",
	publisher = "OTexts",
	address = "Australia",
	edition = "3rd",
}

@article{MAKRIDAKIS:2020,
	title = {{The M4 Competition: 100,000 time series and 61 forecasting methods}},
	journal = {International Journal of Forecasting},
	volume = {36},
	number = {1},
	pages = {54-74},
	year = {2020},
	url = {https://doi.org/10.1016/j.ijforecast.2019.04.014},
	author = {Spyros Makridakis and Evangelos Spiliotis and Vassilios Assimakopoulos},
}

@article {Martinez:2019,
	author = {Mart\'{i}nez, Francisco and Fr\'{i}as, Mar\'{i}a Pilar and P\'{e}rez, Mar\'{i}a Dolores and Rivera, Antonio Jes\'{u}s},
	title = {A methodology for applying k-nearest neighbor to time series forecasting},
	journal = {Artif Intell Rev},
	volume = 52,
	number = 3,
	pages = {2019--2037},
	year = {2019},
	url = {https://doi.org/10.1007/s10462-017-9593-z},
}

@article{Martinez:2022,
	title = {Strategies for time series forecasting with generalized regression neural networks},
	journal = {Neurocomputing},
	volume = {491},
	pages = {509-521},
	year = {2022},
	url = {https://doi.org/10.1016/j.neucom.2021.12.028},
	author = {Francisco Martínez and Francisco Charte and María Pilar Frías and Ana María Martínez-Rodríguez},
}

@ARTICLE{Martinez:2025,
	author={Frías, María P. and Martínez, Francisco},
	journal={IEEE Access}, 
	title={An Ensemble for Automatic Time Series Forecasting With K-Nearest Neighbors}, 
	year={2025},
	volume={13},
	number={},
	pages={4117-4125},
	doi={10.1109/ACCESS.2025.3525561}}

@book{ord2017principles,
	title={Principles of Business Forecasting--2nd Ed},
	author={Ord, J.K. and Fildes, R. and Kourentzes, N.},
	year={2017},
	publisher={wessex, Incorporated}
}

@inproceedings{Q1992,
	author = {J.R. Quinlan},
	title = {Learning with Continuous Classes},
	booktitle = {Proceedings of the 5th Australian Joint Conference on Artificial Intelligence},
	pages = {343--348},
	year = {1992}
}

@article{Taieb:2012,
	author = {Ben Taieb, Souhaib and Bontempi, Gianluca and Atiya, Amir F. and Sorjamaa, Antti},
	title = {A Review and Comparison of Strategies for Multi-step Ahead Time Series Forecasting Based on the {NN5} Forecasting Competition},
	journal = {Expert Syst. Appl.},
	volume = {39},
	number = {8},
	year = {2012},
	pages = {7067--7083},
	url = {https://doi.org/10.1016/j.eswa.2012.01.039}
}
\end{document}